\documentclass{article}
\usepackage[latin1]{inputenc}
\usepackage{times}
\usepackage[TS1,T1]{fontenc}
\newcommand{\tsone}{\fontencoding{TS1}\selectfont}
\usepackage[latin1]{inputenc}
\usepackage[dvips]{graphicx}
\usepackage{latexsym}
\usepackage{xspace}
\usepackage{ml2k}

\usepackage{amsmath}
\usepackage{amsthm}
\usepackage{amssymb} 
\usepackage{algorithm}
\usepackage{algorithmic}
\usepackage{fancyhdr}

\renewcommand{\(}{\left(}
\renewcommand{\)}{\right)}

\renewcommand{\[}{\left[}
\renewcommand{\]}{\right]}

\newcommand{\R}{{\mathbb R}}

\usepackage{color}

\definecolor{darkred}{rgb}{0.7,0.2,0.2}

\definecolor{bgblue}{rgb}{0.04,0.39,0.53}

\begin{document}

\thispagestyle{fancy}
\twocolumn[
\icmltitle{92{\tsone ¢} /MFlops/s, Ultra-Large-Scale Neural-Network Training on a PIII Cluster\\[2mm]
{\small Gordon Bell Price/Performance winner. Student paper award finalist} } 
\icmlauthor{Douglas Aberdeen (corresponding, presenting and student author)}{Douglas.Aberdeen@anu.edu.au}
\icmlauthor{Jonathan Baxter}{Jonathan.Baxter@anu.edu.au}
\icmladdress{Research School of Information Sciences and Engineering, 
Australian National University, Canberra, Australia, 0200}
\icmlauthor{Robert Edwards}{Robert.Edwards@anu.edu.au}
\icmladdress{Department of Computer Science, Australian National University, 
Canberra, Australia, 0200}
]
 
\renewcommand{\floatpagefraction}{0.85}                                         
\renewcommand{\textfraction}{0.05}
\setlength{\textfloatsep}{5pt}
\newcommand{\codename}{{Emmerald}}

\begin{abstract}
Artificial neural networks with millions of adjustable parameters and
a similar number of training examples are a potential solution for
difficult, large-scale pattern recognition problems in areas such as
speech and face recognition, classification of large volumes of web
data, and finance. The bottleneck is that neural network training
involves iterative gradient descent and is extremely computationally
intensive. In this paper we present a technique for distributed
training of {\em Ultra Large Scale Neural Networks}\footnote{Following
the convention with integrated circuits, we take ULSNN to mean a
neural network with in excess of one million parameters and one
million training examples.}  (ULSNN) on {\em Bunyip}, a Linux-based
cluster of 196 Pentium III processors. To illustrate ULSNN training we
describe an experiment in which a neural network with 1.73
million adjustable parameters was trained to recognize
machine-printed Japanese characters from a database containing 9
million training patterns. The training runs with a average performance of
163.3 GFlops/s (single precision). With a machine cost of \$150,913, this
yields a price/performance ratio of 92.4{\tsone ¢} /MFlops/s (single
precision).  For comparison purposes, training using double precision
and the ATLAS DGEMM produces a sustained performance of 70 MFlops/s or
\$2.16 / MFlop/s (double precision).

\end{abstract}
\section{Introduction} 

Artificial neural networks are a class of parametric, non-linear
statistical models that have found wide-spread use in many pattern
recognition domains, including speech recognition, character
recognition, signal processing, medical diagnosis and finance. The
typical network in such an application has 100--100,000 adjustable
parameters and requires a similar number of training patterns in order
to generalize well to unseen test data.  Provided sufficient training
data is available, the accuracy of the network is limited only by its
representational power, which in turn is essentially proportional to
the number of adjustable parameters. Thus, in domains where large
volumes of data can be collected --- such as speech, face and
character recognition, and web page classification --- improved
accuracy can often be obtained by training a much larger network.

In this paper we describe a method for distributed training of {\em
Ultra Large Scale Neural Networks} (ULSNN), or networks with more than
one million adjustable parameters and a similar number of training
examples.  At its core, the algorithm uses {\em Emmerald}, a
single-precision (32 bit) general matrix-matrix multiply (SGEMM) based
on the Pentium III SIMD Streaming Extensions (SSE), with a peak
performance in excess of 1090 MFlops/s on a single 550 MHz Pentium
III. The use of single-precision floating point operations is
justified by the fact that we have found it sufficient for
gradient-based training of ULSNN's. For medium--large scale neural
networks as few as 16 bits precision is sufficient
\cite{asanovic:1991}.

To illustrate the use of our ULSNN training code, we describe an
experiment in which a neural network with 1.73 million adjustable
parameters is being trained to recognize machine-printed Japanese
characters from a database containing 9 million training patterns.
The training is running on {\em Bunyip}, a 196 processor, Linux-based
Intel Pentium III cluster consisting of 98 dual 550 MHz processor
PC's, each containing 384 MBytes of RAM, 13 GBytes of hard disk and
3x100 Mb/s fast Ethernet cards. All components in Bunyip are ``COTS''
(Commodity-Off-The-Shelf), and were sourced from a local PC
manufacturer (see http://tux.anu.edu.au/Projects/Beowulf/).

Our longest experiment took $56$ hours and $52$ minutes, requiring a
total of $31.2$ Peta Flops ($10^{15}$ single-precision floating-point
operations), with an average performance of 152 GFlops/s (single
precision) while under load.  With no other user
processes running the performance increases to 
163.3 GFlops/s which was sustained for
a four hour test before returning access to other users.  Total memory
usage during training was 32.37 GBytes.  The total machine cost,
including the labor cost in construction, was AUD\$253,000, or
USD\$150,913 at the exchange rate of AUD\$1 = .5965{\tsone ¢} USD on
the day of the final and largest payment. This gives a final
price/performance ratio of USD 92.4{\tsone¢} /MFlops/s (single
precision).  For comparison purposes, training using double precision
and the ATLAS DGEMM
\cite{whaley:1997} produced a sustained performance of 70 MFlops/s or
\$2.16 /MFlops/s (double precision).

\section{``Bunyip'' Hardware Details}

The machine used for the experiments in this paper is {\bf ``Bunyip''}, a
98-node, dual Pentium III Beowulf-class system running Linux kernel
2.2.14.  Our main design goals for this machine were to maximise CPU
and network performance for the given budget of AUD \$250,000 (about
USD \$149,125).  Secondary factors to be balanced into the equation
were: amount of memory and disk; reliability; and the overall size of
the machine. All dollar figures quoted in the remainder of this
paper are US dollars.

The Intel Pentium III processors were chosen over Alpha or SPARC processors
for price/performance reasons.
Dual-CPU systems were preferable as overall
cost and size per CPU is lower than single-CPU or quad-CPU
systems.
Unfortunately, at the time of designing this
machine AMD Athlon and Motorola/IBM G4 systems were not available in dual-CPU
configurations. We were also keen to use the SSE floating point
instructions of the Pentium III range. 550 MHz CPUs were eventually selected
as having the best price/performance available in the Pentium III range at
that time.

For the networking requirements, we decided to go with a
commodity solution rather than a proprietary solution. Gigabit ethernet was
considered, but deemed too expensive at around \$300 per node for the NIC
and around \$1800 per node for the switch. Instead, a novel arrangement of
multiple 100 Mb/s NICs was selected with each node having three NICs which
contributed some \$65 per node (plus switch costs -- see 
below).

The configuration for each node is dual Intel Pentium III 550 CPUs on
an EPoX KP6-BS motherboard with 384 MBytes RAM, 13 GByte UDMA66 (IDE) hard disk
and three DEC Tulip compatible 100 Mb/s network interfaces, one of which has
Wake-On-LAN capability and provision for a Boot ROM. The nodes have no
removable media, no video capability and no keyboards. Each node cost 
\$1282.

\begin{figure}
\begin{center}
\includegraphics[scale=0.35]{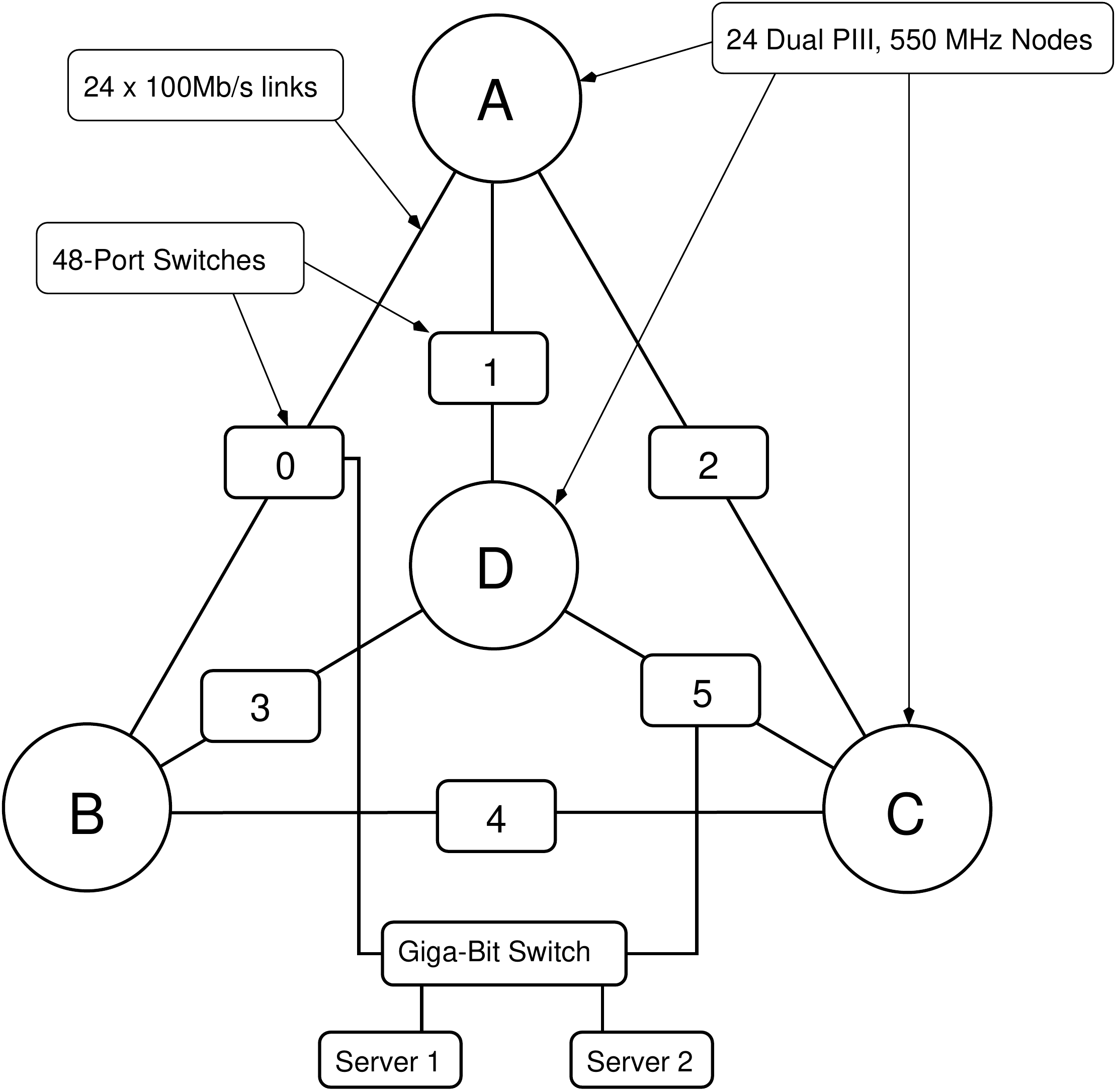}
\caption{Bunyip architecture\label{figure:Bunyip}}
\end{center}
\end{figure}

With reference to figure \ref{figure:Bunyip}, logically the 96 nodes
are connected in four groups of 24 nodes arranged as a tetrahedron
with a group of nodes at each vertex. Each node in a vertex has its
three NICs assigned to one of the three edges emanating from the
vertex.  Each pair of vertices is connected by a 48-port
Hewlett-Packard Procurve 4000 switch (24 ports connecting each way).
The switching capacity of the Procurve switches is 3.8 Gb/s. The
bi-sectional bandwidth of this arrangement can be determined by
looking at the bandwidth between two groups of nodes and the other two
groups through 4 switches, giving a total of 15.2 Gb/s. The 48-port
switches cost
\$2386 each.

Two server machines, more or less identical to the nodes, with the
addition of CD-ROM drives, video cards and keyboards, are each
connected to a Netgear 4-port Gigabit switch which is in turn
connected to two of the HP Procurve switches via gigabit links. The
two server machines also act as connections to the external
network. Two hot-spare nodes were also purchased and are used for
development and diagnostic work when not required as replacements for
broken nodes.

\subsection{Total Cost}
\label{Total Cost}

All up we spent 98 x \$1282 (\$125,636) on the computational 
nodes (including the two hot-spares),
\$17,594 on the six 48-port and the 4-port gigabit switches (6 x \$2386, 2 x
\$894 (gigabit interfaces) 
and \$1490 for the gigabit switch), \$3870 on servers
(including
gigabit NICs, monitors etc.), \$944 for network cables, \$179 on electrical
work,
\$238 on power cables and power boards, and \$298 
on boot EPROMs. The ex-library
shelving was loaned to us, but would have cost \$354 from a local second-hand
furniture shop. Although no component was explicitly budgeted for staff time,
this amounted to about 3 weeks to assemble and configure the machine which
adds approximately \$1800 to the overall cost of the machine. All up, the
total cost was USD \$150,913.

\section{\codename: A SIMD SGEMM for Intel Pentium III Processors}
\label{sgemm}

This section introduces {\em Emmerald}, the high performance software
kernel of our ULSNN training system. It provides a single-precision,
dense, matrix-matrix multiplication routine that uses the single
instruction, multiple data (SIMD) features of Intel PIII chips (SIMD
Streaming Extensions, or SSE). The SSE provide a set of new
floating-point assembler instructions that allow simultaneous
operation on four single-precision floating-point numbers.  Emmerald
outperforms a naive (3-loop) matrix-matrix multiply by 8 times for
square matrices of size $64$, and a peak of 29 times for matrices of
size $672$.  Emmerald can be downloaded from
http://beaker.anu.edu.au/$\sim$daa/research.html.

\subsection{Single precision general matrix-matrix multiply (SGEMM)}

Without resorting to the complexities associated with implementing
Strassen's algorithm on deep-memory hierarchy machines
\cite{strassen:1969,thottethodi:1998}, dense matrix-matrix
multiplication requires $2MNK$ floating point operations where $A:M
\times K$ and $B:K \times N$ define the dimensions of the two
matrices. Although this complexity is fixed, skillful use of the
memory hierarchy can dramatically reduce overheads not directly
associated with floating point operations. 
Memory hierarchy optimization combined with the use of SSE gives 
Emmerald its performance
advantage. 

Emmerald implements the SGEMM interface of Level-3 BLAS, and so may be
used to improve the performance of single-precision
libraries based on BLAS (such as LAPACK \cite{blas:1998}).  
There have been several recent attempts at automatic optimization of
GEMM for deep-memory hierarchy machines, most notable are
PHiPAC \cite{bilmes:1996} and the more recent ATLAS \cite{whaley:1997}. 
ATLAS  in particular achieves performance close to 
vendor optimized commercial GEMMs. Neither ATLAS nor PhiPAC make use
if the SSE instructions on the PIII for their implementation of
SGEMM. 

\subsection{SIMD Parallelisation}
\label{SIMD Parallelisation}

\begin{figure}
\begin{center}
\includegraphics[scale=0.8]{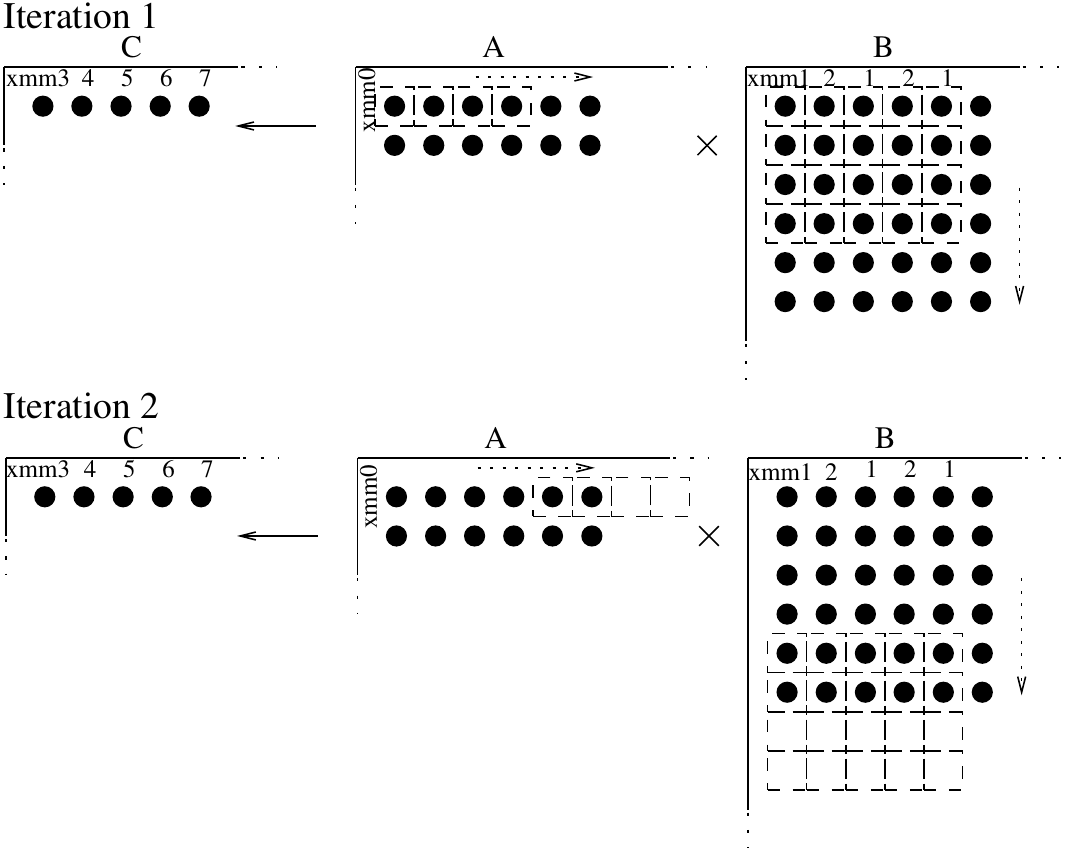}
\caption{Allocation of SSE registers (labelled as \texttt{xmm[0-7]}), 
showing progression of the dot
products which form the innermost loop of the algorithm. Each black
circle represents an element in the matrix.  Each dashed square
represents one floating point value in a SSE register. Thus four
dotted squares together form one 128-bit SSE register.}
\label{f:iter}
\end{center}
\end{figure}

A SIMD GEMM must aim to minimize the ratio of memory accesses to
floating point operations.  We employed two core strategies to achieve
this:
\begin{itemize}
\item accumulate results in registers for as long as possible 
to reduce write backs;
\item re-use values in registers as much as possible.
\end{itemize}
In \cite{greer:1997} several dot-products were performed in parallel
inside the innermost loop of the GEMM. Taking the same approach we
found experimentally that 5 dot-products in the inner loop gave the
best performance.  Figure \ref{f:iter} shows how these 5 dot products
utilise SIMD parallelism.

\subsection{Optimizations}

\begin{figure}
\begin{center}
\includegraphics[scale=0.46]{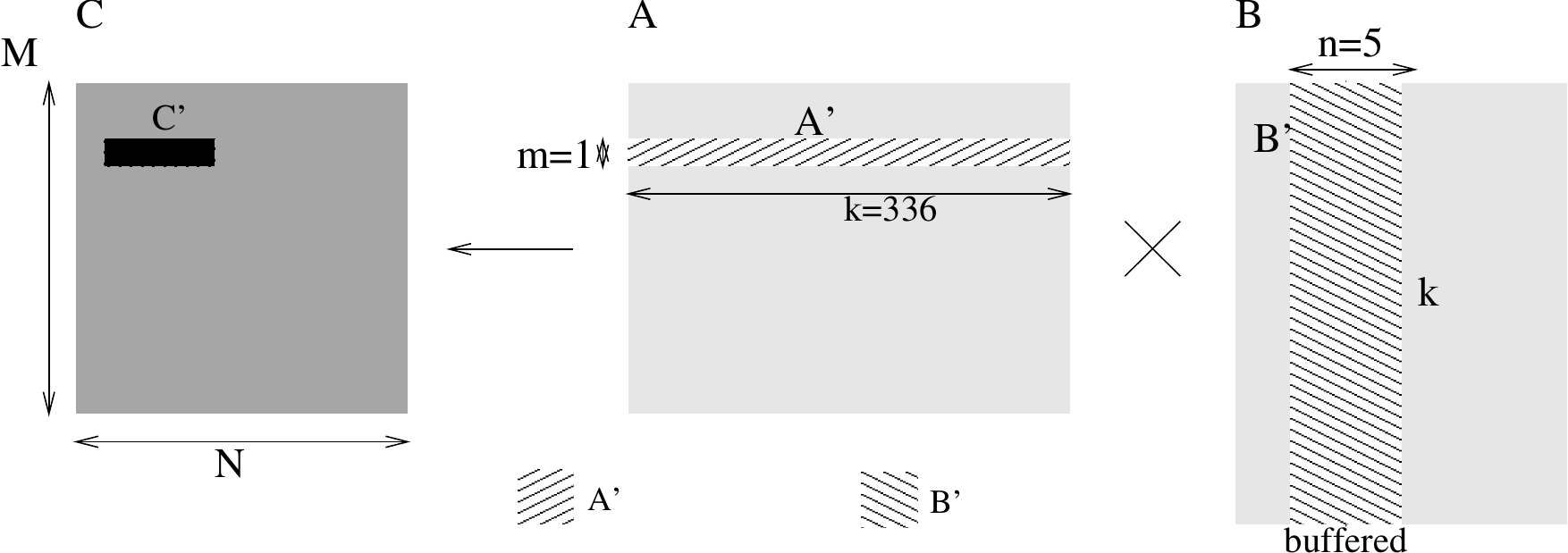}
\caption{L1 blocking for \codename: $C' \leftarrow A'B'$ where $A'$ and $B'$
are in L1 and $C'$ is accumulated in registers.}
\label{f:l1cache}
\end{center}
\end{figure}

A number of techniques are used in \codename\ to improve
performance. Briefly, they include:
\begin{itemize}
\item \emph{L1 blocking}:
\codename\ uses matrix blocking \cite{greer:1997,bilmes:1996,whaley:1997} to
ensure the inner loop is operating on data in L1 cache.
Figure~\ref{f:l1cache} shows the L1 blocking scheme. The block dimensions
$m$ and $n$ are determined by the configuration of dot-products in the 
inner loop 
(Section~\ref{SIMD Parallelisation}) and $k$ was determined experimentally.
\item \emph{Unrolling}: The innermost loop is completely unrolled 
for all possible lengths of 
$k$ in L1 cache blocks, taking care to avoid overflowing the 
instruction cache. 
\item \emph{Re-buffering}:
Since $B'$ (Figure \ref{f:l1cache}) is large $(336 \times 5)$ compared to
$A'$ 
$(1 \times 336)$, we deliberately buffer $B'$ into L1 cache. 
While buffering $B'$ we re-order its elements to enforce 
optimal memory access patterns. This has the additional benefit of
minimising translation look-aside buffer misses \cite{whaley:2000}.
\item \emph{Pre-fetching}: Values from $A'$ are not
buffered into L1 cache. We make use of SSE pre-fetch assembler instructions  
to ensure $A'$ values will be in L1 cache when needed.
\item \emph{L2 Blocking}: Efficient L2
cache blocking ensures that peak rates can be maintained as 
long as $A$, $B$ and $C$ fit into main memory.
\end{itemize}

\subsection{Emmerald Results}

The performance of \codename\ was measured by timing matrix multiply
calls with size $M=N=K=16$ up to 700. 
The following steps were taken to ensure a conservative performance
estimate:
\begin{itemize}
	\item wall clock time on an unloaded machine is used rather than
	 CPU time;
     \item the stride of the matrices, which determines the separation in 
	   memory between each row of matrix data, 
	 is fixed to 700
	rather than the optimal value (the length of the row);
     \item caches are flushed between calls to \texttt{sgemm()}.
\end{itemize}
Timings were performed on a PIII 450MHz running Linux (kernel 2.2.14).

\begin{figure}
\begin{center}
\includegraphics[scale=0.665]{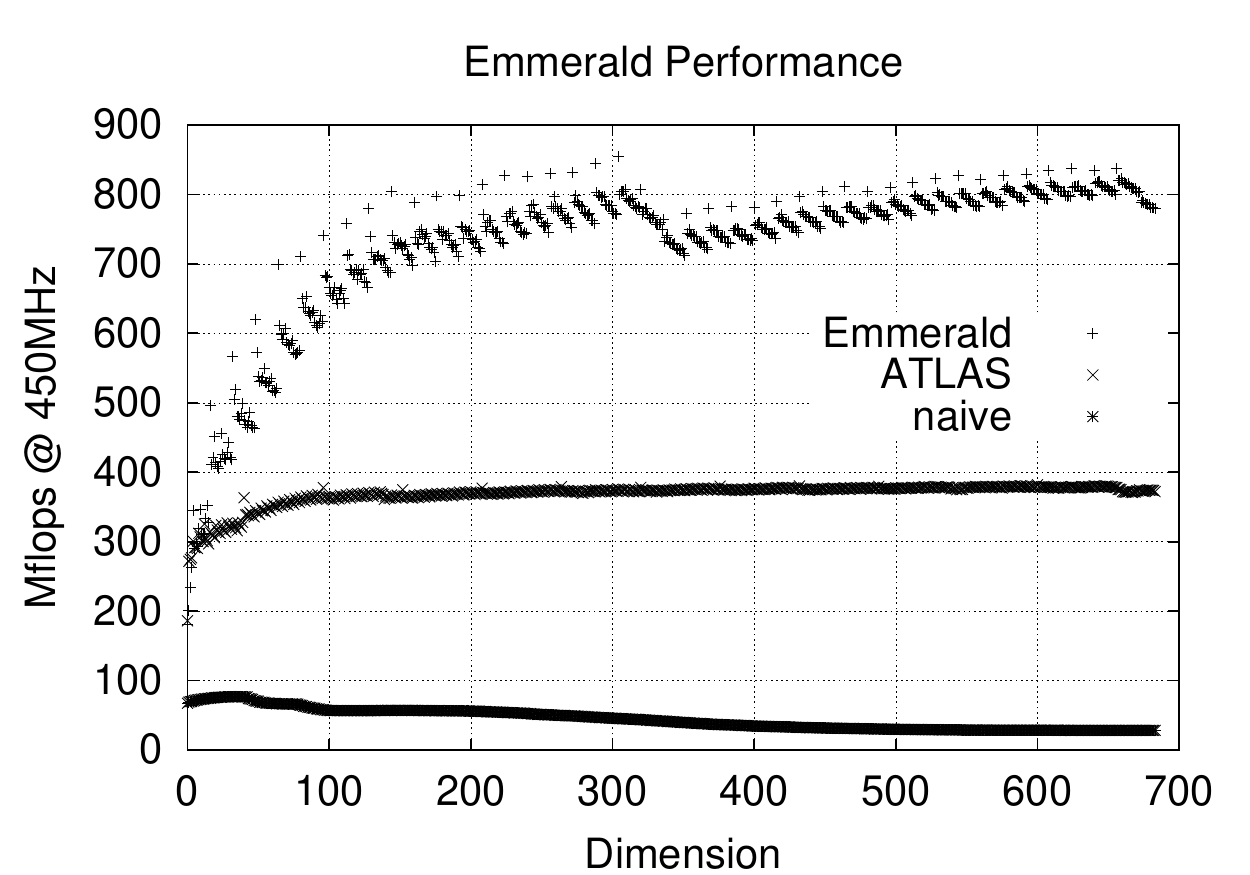}
\caption{Performance of Emmerald on a PIII running at 450MHz compared
to ATLAS sgemm and a naive 3-loop matrix multiply. Note that ATLAS does not
make use of the PIII SSE instructions.}
\label{f:results}
\end{center}
\end{figure}

Figure~\ref{f:results} shows {\codename}'s performance compared to
ATLAS and a naive three-loop matrix multiply. 
The average MFlops/s rate of \codename\ after size
100 is 1.69 times the clock rate of the processor 
and 2.09 times faster than ATLAS. 
A peak rate of 890 MFlops/s is achieved when $m=n=k=stride=320$. 
This represents 1.98 times the clock rate. On a PIII 550 MHz (the
processors in Bunyip) we achieve a peak of 1090 MFlops/s.
The largest tested size was $m=n=k=stride=3696$ which ran at 940 MFlops/s at 
550 MHz. 
For more detail see \cite{aberdeen:1999}.

\section{Training Neural Networks using SGEMM} 
In this section we describe one-hidden-layer artificial neural
networks and, following \cite{bilmes:1997}, how to  compute the
gradient of a neural network's error using matrix-matrix
multiplication. We then describe our conjugate-gradient approach to
training neural networks.

\subsection{Artificial Neural Networks}
A one-hidden-layer artificial neural
network maps input vectors $x=(x_1, \dots, x_{n_i})\in\R^{n_i}$ to output
vectors $y=(y_1, \dots, y_{n_o})\in\R^{n_o}$ according to the formula: 
\begin{equation}
\label{eq:nn}
y_i(x) = \sigma\(\sum_{j=1}^{n_h} w^{ho}_{ij} h_j(x)\),
\end{equation}
where $\sigma:\R \to \R$ is some squashing function (we use $\sigma =
\tanh$), $w^{ho}_{ij}$ are the adjustable parameters connecting the
hidden nodes to the output nodes, and  $h_j(x)$ is the activation of the $j$-th hidden node: 
\begin{equation}
\label{eq:hidden}
h_j(x)  = \sigma\(\sum_{k=1}^{n_i} w^{ih}_{jk} x_k\). 
\end{equation}
In the last expression, $w^{ih}_{jk}$ are the adjustable parameters
connecting the input nodes to the nodes in the hidden layer. Given a
matrix $X$ of $n_p$ training patterns and a matrix $T$ of desired
outputs for the patterns in $X$, 
$$
X = 
\begin{bmatrix} 
x_{11} & \dots & x_{1n_i} \\
\vdots & \ddots & \vdots \\
x_{n_p1} & \dots & x_{n_pn_i}
\end{bmatrix},
T = 
\begin{bmatrix} 
t_{11} & \dots & t_{1n_o} \\
\vdots & \ddots & \vdots \\
t_{n_pn_o} & \dots & t_{n_pn_o}
\end{bmatrix},
$$
the goal is to find sets of parameters $w^{ho}_{ij}$ and
$w^{ih}_{kl}$ minimizing the {\em mean-squared error}: 
\begin{equation}
\label{eq:error}
E = \sum_{i=1}^{n_p} \sum_{j=1}^{n_o} \[y_j(x_i)  - t_{ij}\]^2,  
\end{equation}
where $x_i$ is the $i$-th row of the data matrix $X$. 
Usually \eqref{eq:error} is minimized by some form of gradient
descent. 

\subsection{Computing The Gradient Using Matrix-Matrix Multiply}
If we write $Y$ for the matrix of outputs $y_j(x_i)$, $H$ for
the matrix of hidden activations $h_j(x_i)$, and
$W^{ih}$ and $W^{ho}$ for the parameter matrices $w^{ih}_{kl}$ and
$w^{ho}_{ij}$ respectively, then 
\begin{align*}
H &= \sigma\(X * W_{ih}^T \) \\
Y &= \sigma\( H * W_{ho}^T\) \\
\end{align*}
where ``$*$'' denotes ordinary matrix multiplication
and $\sigma(A)$ means apply $\sigma$ elementwize to the components of
$A$. Defining 
\begin{align*}
Y_\Delta &= \(I - Y *\!* Y\) *\!* \(T - Y\), \\ 
H_\Delta &= \(I - H *\!* H\) *\!* \(Y_\Delta * W_{ho}\), 
\end{align*}
where ``$**$'' denotes elementwize matrix multiplication, we have
\begin{align*}
\nabla_{\!ih} E &= H_\Delta^T * X, \\
\nabla_{\!oh} E &= Y_\Delta^T * H, 
\end{align*}
where $\nabla_{\!ih} E$ is the gradient of $E$ with respect to the
parameters $W_{ih}$ and $\nabla_{\!ho} E$ is the gradient of $E$ with
respect to $W_{ho}$ \cite{bilmes:1997}. 

Thus, computing the gradient of the error for an artificial neural
network can be reduced to a series of ordinary matrix multiplications
and elementwize matrix multiplications. For large networks and large
numbers of training patterns, the bottleneck is the ordinary matrix
multiplications, which we implement using Emmerald's SGEMM routine. In
all our experiments we found 32 bits of floating-point precision were
enough for training. For neural networks with $\approx$ 10,000
parameters, as few as 16 bits are sufficient \cite{asanovic:1991}.

Armed with the gradient $\nabla E$, we can adjust the parameters $W$
by a small amount in the negative gradient direction $W := W - \alpha
\nabla E$ and hence reduce the error. However, because the gradient
computation can be very time-consuming (a total of 52.2 Tera-floating
point operations in our largest experiment), it is more efficient to
employ some form of line search to locate a local maximum in the
direction $\nabla E$. For the experiments reported in the next section
we used the Polak-Ribi\'{e}re conjugate-gradient descent method
\cite[\S5.5.2]{fine99} to choose the search direction, combined with
an exponential step-size scheme and quadratic interpolation in order
to locate a maximum in the search direction. We were also able to
speed the search for a maximum by using gradient information to
bracket the maximum, since only the {\em sign} of the inner product of the
gradient with the search direction is required to locate the maximum in that direction, and 
the sign can be reliably estimated with far fewer training
patterns than is required to estimate the error. 

\subsection{Training Set Parallelism}
Since the error $E$ and gradient $\nabla E$ are {\em additive} over
the training examples, the simplest way to parallelize the training of
a neural network is to partition the training data into disjoint
subsets and have each processor compute the error and gradient for its
subset. This works particularly well if there are a large number of
training patterns so that each processor can work with near-optimal
matrix sizes. The communication required is the transmission of the
neural network parameters to each slave processor, and the
transmission of the error and gradient information back from each
slave to a master node which reduces them to a single error or
gradient vector.

\section{Communication}

This section discusses the communication costs associated with distributed 
NN training, arguing that these costs are non-trivial for ULSNNs. 
A reduce algorithm optimised for Bunyip's topology is also discussed.

\subsection{Communication Costs}
\label{Communication Costs}

The inter--process communication costs during network training 
arise from \emph{broadcasting} the network parameters to all processes
and \emph{reducing} the network error and gradients from each
process to the master process. The parameter broadcasting is cheap, since
many copies of the same data is sent to all processes. Broadcasts can
take advantage of features such as TCP/IP broadcasting. The reduce process
is more difficult with each process generating unique vectors 
which must be collected and summed
by the master process. The time taken to reduce data 
grows with both the number of parameters
and the number of processes. The remaining communication consists of
start and stop messages which are insignificant compared to the aforementioned
costs.
 
A typical neural network with 100 inputs, 50
hidden layer neurons, and 50 output neurons, requires 7500 parameters,
or 30 KBytes of data (single precision), to be sent from every 
node to the master node. A naive
reduction over 194 processes using a 1Gb/s link, such as 
used in Bunyip, would take 0.05 seconds assuming 100\% network utilisation. 
Our ULSNN with 400 inputs, 480 hidden 
layer neurons and 3203 output neurons requires 
1,729,440 parameters or 6.6 MBytes of data per process
which would require 10.1 seconds. There is sufficient memory on each node to
occupy both processors for 446 seconds calculating gradients before a
reduce operation is required. 
Consequently the reduce operation would cost at least 2.3\% of the available 
processing time, more if not enough training data is available 
or the network size is increased.

This demonstrates that although communication costs for 
distributed NN training are
minimal for commonly implemented network sizes, ULSNN training must optimise 
inter--process communication to achieve the best performance.

We reduced communication as much as possible by only distributing the
neural-network parameters to all the slaves at the very start of
training (rather than at each step), and thereafter communicating only the
search direction and the amount to step in that
direction. One significant reduce operation is required per epoch to  
send  the error gradient vector from each process 
to the master which then co-ordinates the step size search with the slaves.

All communication was done using the LAM implementation of
MPI (http://www.mpi.nd.edu/lam). Communicating parameters or directions to all
processors required a 6.6 MBytes broadcast operation from the server to
each of the 194 processors in the cluster, while reducing the gradient back
to the master required 6.6 MBytes of data to be communicated from each
processor back to the server. LAM/MPI contains a library reduce operation which
uses a simple $O(\log n)$ algorithm that distributes the load of the reduce
over many processes instead of naively sending 194 gradient vectors to one
node \cite{lamcode}. 
This results in a reduce operation on Bunyip which takes 
8.5 seconds over 8 stages.

\subsection{Optimising Reductions}

There are two problems with existing free implementations of MPI reduce 
operations. 
The first is the lack of shared memory protocols
on clusters with multi-processor nodes, instead 
using slow TCP/IP communications between
processors on the same motherboard. Secondly, the reduce operation
does not take advantage of the topology of the cluster. For example, the
best reduce algorithm to use on a ring network might be to send a single vector
to each node on the ring in turn, which adds its contribution before 
passing the vector
to the next node. On a star network the best algorithm might be to send
each contribution to the central server and sum as they arrive.

\begin{figure}
\begin{center}
\includegraphics[scale=0.6]{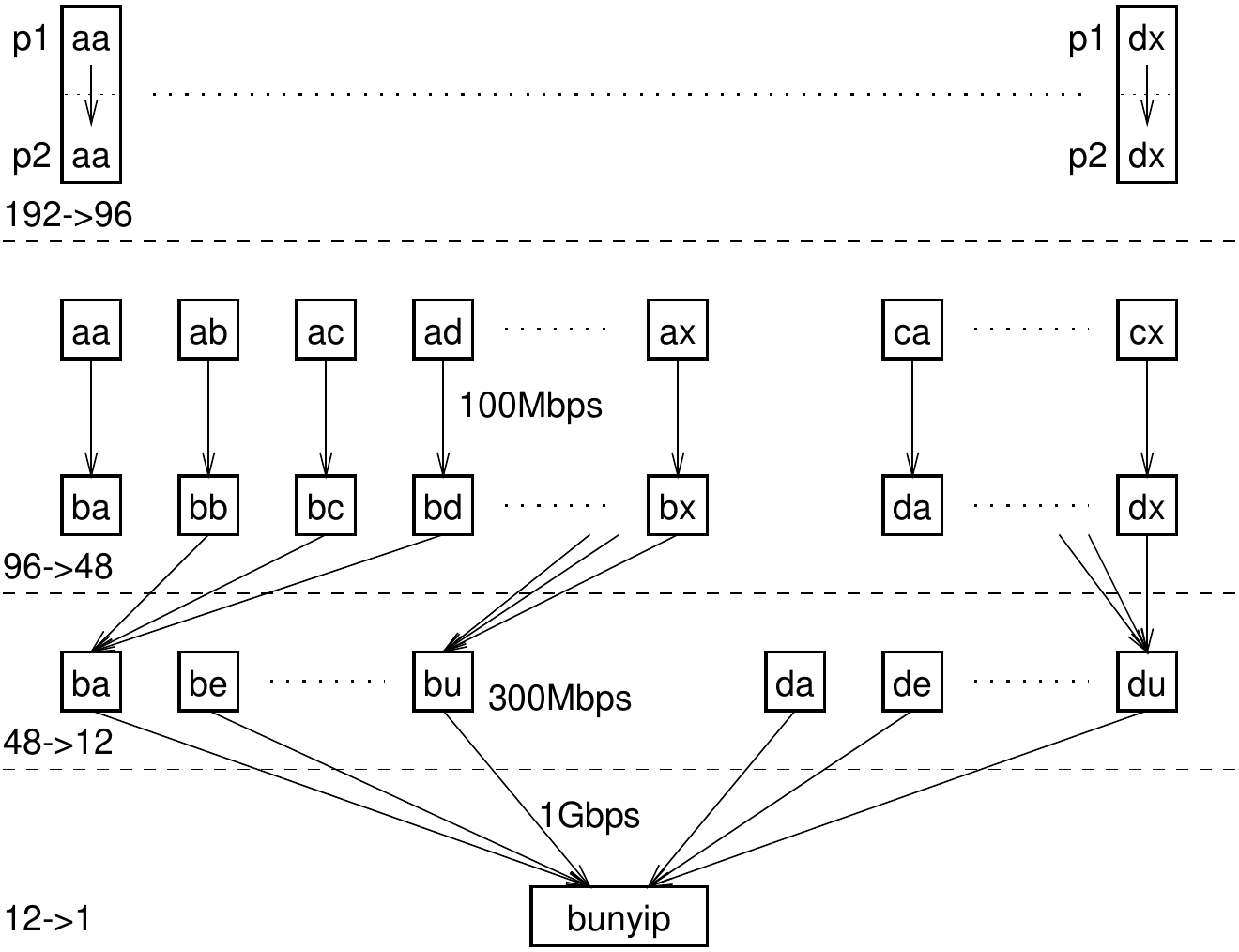} 
\caption{The four stages of our customized reduce: Stage 1: SHM intra-node 
reduce; stage 2: all nodes in group A and C reduce to their
counterparts; stage 3: groups B and D reduce to 12 nodes using 3 NICs;
stage 4: MPI library reduce to the server node.}
\label{f:reduce}
\end{center}
\end{figure}

To decrease the time taken per reduce, 
we wrote a customized
routine utilising shared memory for intra-node communication 
and MPI non-blocking calls for inter-node communication. This routine
is summarised by Figure~\ref{f:reduce}. It is split into 4 stages, each of
which takes advantage of an aspect of Bunyip's 
topology shown in Figure~\ref{figure:Bunyip}.
\begin{enumerate}
\item 
	Each node contains two processors, both running an instance of the 
	training process. All 97 nodes (including the server), 
	reduce 6.6 MBytes of data though shared memory between processes, 
	taking 0.18 seconds. The time taken to add the two sets of data
	together is approximately 0.005 seconds.

\item

	Each node in group A can open a 100 Mb/s connection to any
	node in group B via switch 0. Thus all 24 nodes in A can
	reduce to their B counterparts in parallel.  This requires
	0.66 seconds. The same trick is used for reducing from group C to D.
	The reduced data now resides only on the B and D
	nodes. The total bandwidth for all 96 nodes in this stage is
	4.03 Gb/s.

\item 

	Each node contains 3x100 Mb/s NICs. This allows a node to receive 
	data from three other nodes simultaneously provided the TCP/IP 
	routing tables are correctly configured. We split the 24 nodes
	in each group into 6 sets of 4 nodes. The first of each set
	(see node BA in Figure~\ref{f:reduce}) is designated as the
	root and the other three nodes send to it via different NICs.
	This takes 0.9 seconds achieving a bandwidth of 185 Mb/s into
	each root node, or 2.22 Gb/s across all 12 root nodes. 

\item 

	The final step is a standard MPI library reduce from 6 B nodes and
  	6 D nodes to the master process. This is the slowest step
	in the process taking 3.16 seconds, including the time spent waiting
	for the the nodes to synchronize since they do not start reducing
	simultaneously.

\end{enumerate}

\begin{figure}
\begin{center}
\includegraphics[scale=0.665]{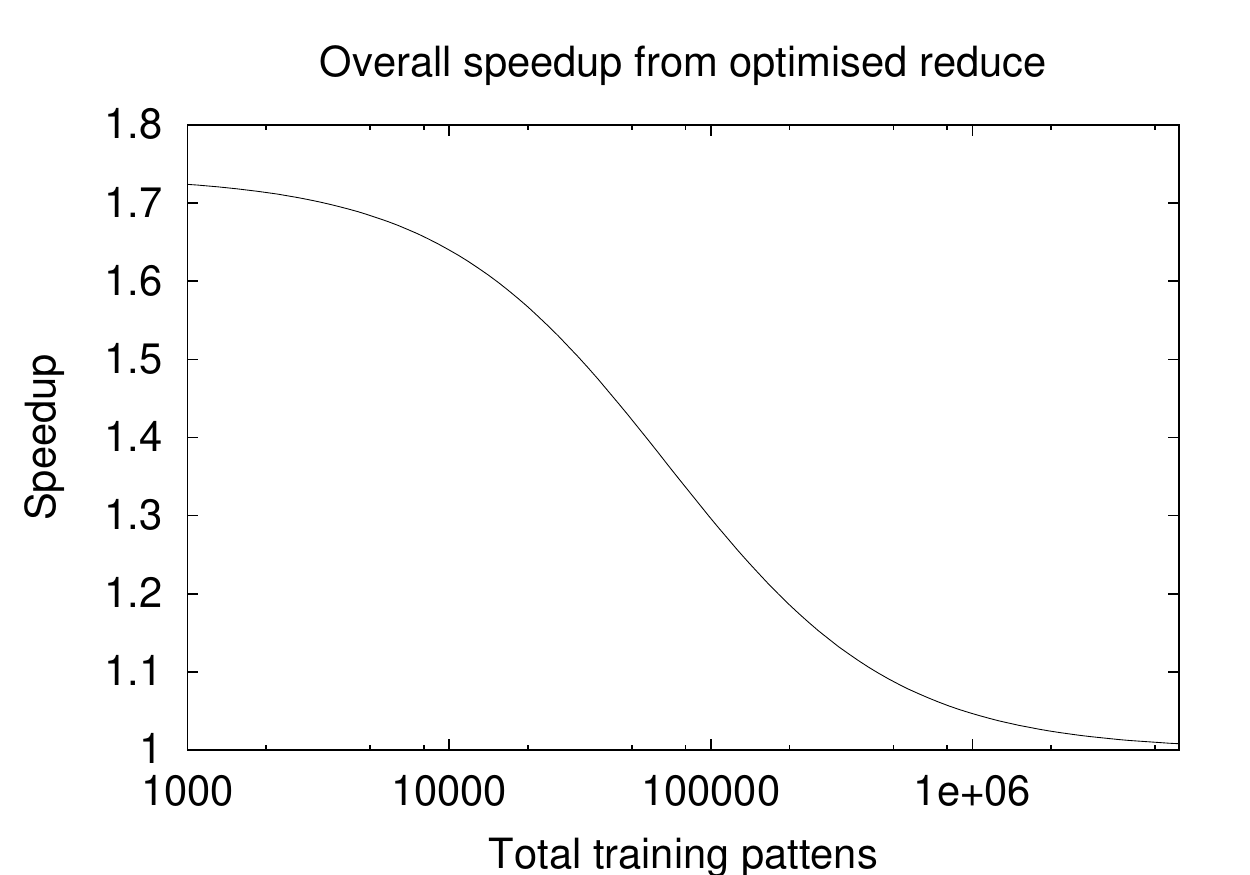} 
\caption{The overall training performance speedup exhibited after replacing
the MPI library reduce with our optimised reduce against the total
number of training patters used.}
\label{f:reducespeed}
\end{center}
\end{figure}

The overall time taken for the optimised reduce to complete is 4.9
seconds.  The actual time saved per reduction is 3.6 seconds.  The
training performance speedup from this saving varies with the duration
of the gradient calculation which depends linearly on the number of
training patterns.  Figure~\ref{f:reducespeed} illustrates the
expected speedup achieved by using the optimised reduce instead of
the MPI library reduce, against the total number of training patterns
used. In practice our peak performance of 163.3 GFlops/s benefits by
roughly 1\% from the optimised reduce, however the speedups are much
more marked for smaller (and more frequently encountered) data sets.

\section{Japanese Optical Character Recognition} 

In this section we describe our distributed application of 
the matrix-matrix multiply technique of Section~\ref{sgemm} used
to train an artificial neural network as a classifier for
machine-printed Japanese characters.

\subsection{The Problem, Data and Network Architecture} 
Japanese optical character recognition (Japanese OCR) is the process
of automatically recognizing machine-printed Japanese documents and
converting them to an electronic form. The most difficult aspect of
Japanese OCR is correctly classifying individual characters, since
there are approximately 4000 characters in common usage.

\begin{figure}
\begin{center}
\includegraphics{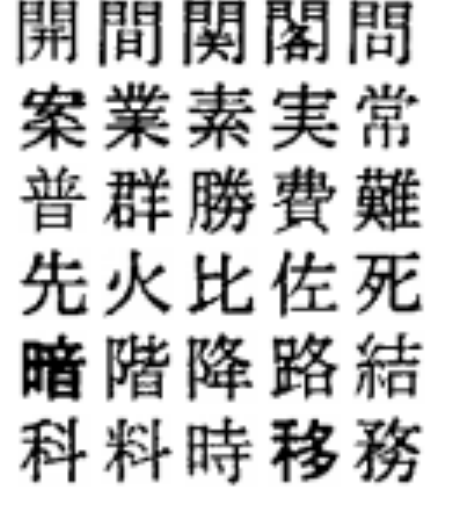} 
\caption{Example Japanese characters used to train the neural network.\label{figure:jchars}}
\end{center}
\end{figure}

The training data for our neural network consisted of 168,000
scanned, segmented, hand-truthed images of Japanese characters
purchased from the CEDAR group at the University of Buffalo. The
characters were scanned from a variety of sources, including books,
faxes, newspapers and magazines. Figure \ref{figure:jchars} gives an
idea of the varying quality of the character images. 

Each character in the CEDAR database is represented as a binary image
of varying resolution. We down-sampled all the images to a $20 \times
20$ grey-scale format. The neural network had $400$ input nodes, one
for each pixel. The database contained examples of $3203$ distinct
characters, hence the neural-network had $3203$ output nodes.  The
hidden layer was chosen to have $480$ nodes. In total, the network had
$1.73$ million adjustable parameters.

168,000 training examples are not sufficient to avoid overfitting in a
network containing 1.73 million adjustable parameters, so we generated
synthetic data from the original characters by applying random
transformations including line thickening and thinning, shifting,
blurring and noise addition. The total number of training examples
including the artificial ones was 9,264,000 approximately $5.4$ per
adjustable network parameter. These were distributed uniformly to 193
of the processors in Bunyip.  A further 6320 examples of the CEDAR
data set were used for testing purposes.

\subsection{Training}

With reference to equations \eqref{eq:nn}, \eqref{eq:hidden}, and
\eqref{eq:error}, the total number of floating point operations required to compute the error
$E$ in a neural network is $2 \times n_p \times (n_i + n_o) \times
n_h$, which equals $32$ Tera floating-point operations for the
Japanese OCR experiment. A gradient calculation uses $n_p
\times (4 \times n_i \times n_h + 6 \times n_h \times n_o)$, or $92$
Tera floating-point operations. 

To assist with load balancing, each slave processor stepped through its
training patterns $320$ at a time. Between each step the master node
was polled to determine whether more steps were required. Once $80\%$
of the total training data had been consumed, the master instructed
all slaves to halt computation and return their results (either the
error or the gradient). In this way the idle time spent
waiting for other slaves to finish was reduced to at most the length of time
needed by a single processor to process $320$ patterns. With $80\%$ of
the data, an error calculation required $26$ TFlops and a
gradient calculation requires $74$ TFlops, or $135$ GFlops and
$383$ GFlops per processor respectively.

\section{Results} 

This section describes the classification accuracy achieved; then
concentrates on the performance scalability over processors before
finishing with peak performance results which result in our claim of a
price/performance ratio of 92.4{\tsone ¢} /MFlop/s.
 
\subsection{Classification Accuracy}

The network's best classification error on the held-out 6,320
examples is 33\%, indicating substantial progress on a difficult problem 
(an untrained classifier has an error of  $1 - 1/3200 = 99.97\%$). 
We observed an error rate of 5\% on the 40\% of the data which
contained the most examples of individual characters. Continued training after
the 33\% error rate was achieved improved the performance on the common
characters at the cost of greatly decreased performance on the rare ones.
This leads to the conclusion that overfitting is occurring on characters
with only one or two examples from the original data set, despite the
number of transformations being generated.
A more uniform accuracy could be achieved
by generating more transforms of rare characters, or preferably,  using a 
greater number of original examples.

A very large amount of data is required for two reasons. The first is to 
avoid overfitting. Table~\ref{t:genperf} compares the generalisation
accuracy with the total number of training examples used (including 
transformations of the original 168,000 patterns). Each data point in
this graph represents approximately 48 hours training time. Training
was halted after 10 epochs result in no classification improvement on
the test set.

\begin{table}
\begin{center}
\begin{tabular}{|r|r|}
\hline
Patterns & \% Error\\
\hline 
343800 & 51 \\
611200 & 46 \\
1833600 & 33\\
\hline
\end{tabular}
\caption{Generalisation error decreases as the total number of patterns
increases.}
\label{t:genperf}
\end{center}
\end{table}

\subsection{Communication Performance}

\begin{figure}
\begin{center}
\includegraphics[scale=0.665]{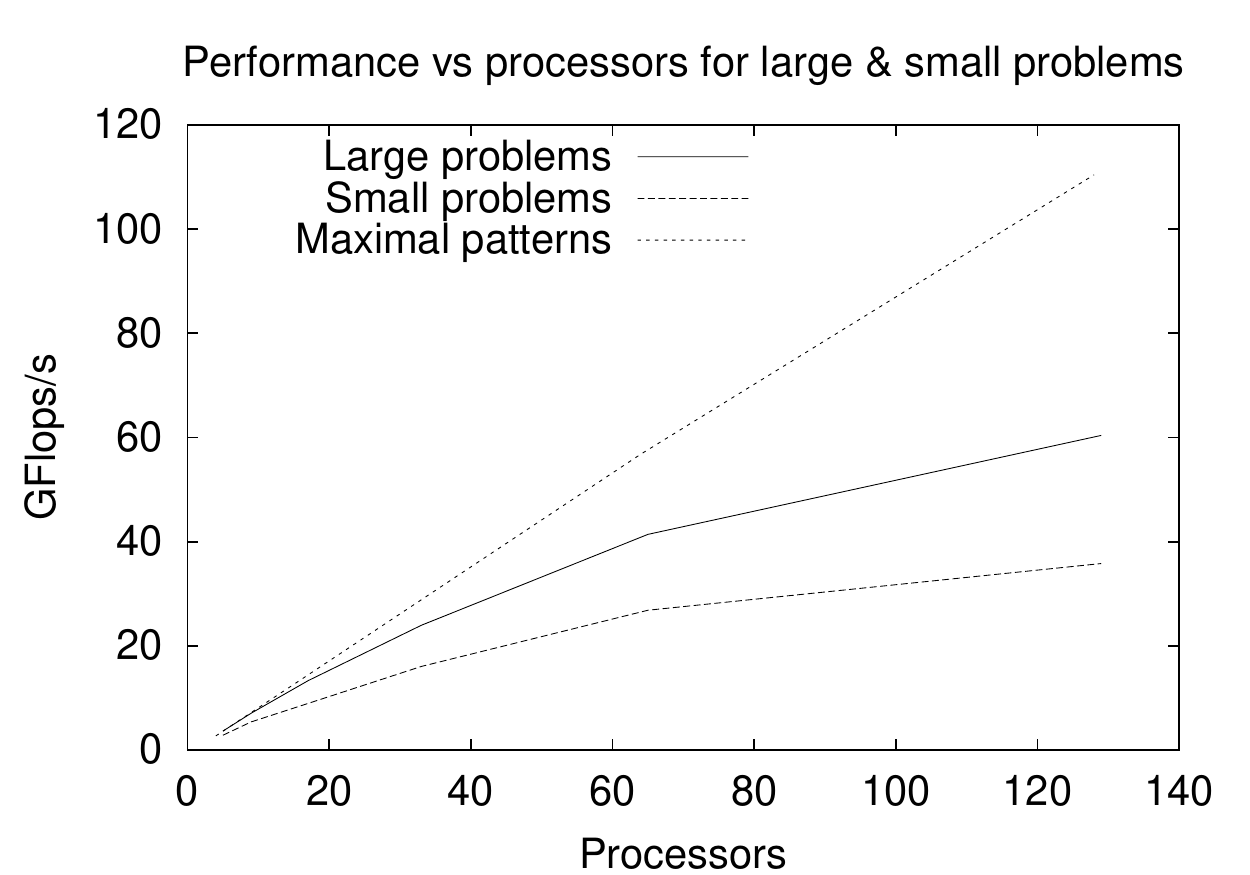}
\caption{Performance scaling with the number of processors used for
training a small network and our large JOCR network with a fixed number 
of patterns, and the JOCR problem when the total patterns scales
with the number of processors.}
\label{f:procscale}
\end{center}
\end{figure}

Recalling from Section~\ref{Communication Costs} that communication
overhead increases with decreasing patterns then the second motivation
for large training sets is to reduce such
overhead. Figure~\ref{f:procscale} demonstrates how the performance
scales with the number of processors used.  The bottom line is the
performance versus processors curve for a small network of 400 input
nodes, 80 hidden layer nodes, 200 output nodes and a total of 40,960
training patterns. The middle line is our JOCR ULSNN with 163,480 total
patterns. The top line is the JOCR network again, however, for this test
we allowed the number of patterns to scale with the processors,
minimizing the frequency of reduce operations. The maximal patterns test
uses 32,000 patterns per processor. 
All performance values quoted in this paper represent the
total flops that contribute to feed forward value and gradient
calculations divided by the wall clock time. Implementation specific
flops, such as the reduce operations, were not included.
Bunyip was under a small
load during the performance testing for Figure~\ref{f:procscale}.

For a small number of processors, both networks exhibit linear
performance scale up, but we observe that for many processors 
the larger problem scales better despite the increased number 
of network parameters. This
is due to the communication overhead in the small network increasing
dramatically as each processor has less data to process before needing
to initiate a reduce. The effect would be clearer for a large network
(causing long gradient vectors to be reduced) with few training patterns, 
however this scenario is not usually encountered due to overfitting.
Finally we observe that with a large enough data set to fill 
the memory of every node, we achieve near linear scaling.

\subsection{Price/Performance Ratio}

Bunyip was dedicated to running the 
JOCR problem for four hours with 9,360,000
patterns distributed across 196 processors. Bunyip actually consists
of 194 processors, however, we co-opted one of the hot-spare nodes
(included in the quoted price) to make up the other two processors.

Over this four hour period a total of 2.35 PFlops were performed with an
average performance of 163.3 GFlops/s. This performance is
sustainable indefinitely provided no other processes use the
machine. To calculate the price/performance ratio we use the total
cost derived in Section~\ref{Total Cost} of USD\$150,913, which yields 
a ratio of 92.4{\tsone ¢}~/MFlop/s\footnote{Using the
exchange rate at the time of writing would yield 91.5{\tsone¢} USD /MFlops/s,
however we felt quoting the rate at the time of purchase was a more
accurate representation of the cost.}.

\section{Conclusion} 

We have shown how a COTS (Commodity-Off-The-Shelf) Linux Pentium III
cluster costing under
\$151,000 can be used to achieve sustained, Ultra-Large-Scale
Neural-Network training at a performance in excess of 160 GFlops/s
(single precision), for a price/performance ratio of 92.4{\tsone ¢}/MFlop/s.

Part of the reason for the strong performance is the use of very large
training sets. With the current networking set-up, performance
degrades significantly with less data per processor, as communication
of gradient information starts to dominate over the computation of the
gradient.

\section*{Acknowledgements}

This project was supported by the Australian Research Council, an
Australian National University Major Equipment Grant, and LinuxCare
Australia. Thanks are also due to several people who made valuable
contributions to the establishment and installation of Bunyip: Peter
Christen, Chris Johnson, John Lloyd, Paul McKerras, Peter Strazdins
and Andrew Tridgell.

\small
\bibliographystyle{abbrv}
\bibliography{Bib/bib}

\end{document}